\relax

\documentclass[letterpaper]{article} 
\usepackage[switch]{lineno}  %
\usepackage{aaai21}  
\usepackage{times}  
\usepackage{helvet} 
\usepackage{courier}  
\usepackage[hyphens]{url}  
\usepackage{graphicx} 
\usepackage{graphicx}  
\usepackage{natbib}  
\usepackage{caption} 
\usepackage{amsmath}
\usepackage{amsfonts}
\usepackage{tabularx}
\usepackage{booktabs}

\usepackage{hyperref}
\usepackage{cleveref}

\DeclareMathOperator*{\argmin}{argmin}
\title{A Context Integrated Relational Spatio-Temporal Model for Demand and Supply Forecasting}
\author{
    Hongjie Chen,\textsuperscript{\rm 1}
    Ryan A. Rossi,\textsuperscript{\rm 2}
    Kanak Mahadik,\textsuperscript{\rm 2}
    Hoda Eldardiry\textsuperscript{\rm 1}
    \\
}
\affiliations{
    \textsuperscript{\rm 1}Department of Computer Science, Virginia Tech\\
    Blacksburg, Virginia 24060\\
    \textsuperscript{\rm 2}Adobe Research\\
    San Jose, California 95110\\
    jeffchan@vt.edu, rrossi@adobe.com, mahadik@adobe.com, hdardiry@vt.edu\\
}

\usepackage{algorithmic}
\usepackage{amsmath,amssymb,amsfonts}
\usepackage{amssymb}
\usepackage{booktabs}
\usepackage{comment}
\usepackage{faktor}
\usepackage{graphicx}
\usepackage{mathcomp}
\usepackage{multirow}
\usepackage{pifont}
\usepackage{tabularx}
\usepackage{textcomp}
\usepackage{xcolor}
\usepackage{xparse}
\def\eg{e.g.,~} 
\def\ie{i.e.,~} 
\newtheorem{definition}{Definition}

\newcommand{\figref}[1]{Fig.~\ref{fig:#1}}
\newcommand{\tabref}[1]{Table~\ref{tab:#1}}

\newcommand{\defref}[1]{Definition~\ref{def:#1}}

\newcommand{\tb}[1]{\textbf{#1}}
\newcommand{\ti}[1]{\textit{#1}}
\newcommand{\mb}[1]{\mathbf{#1}}
\newcommand{\card}[1]{\left|{#1}\right|}
\newcommand{\Paragraph}[1]{\noindent\textbf{#1}}

\def\mK{\mathbb{K}}
\def\mR{\mathbb{R}}
\def\mX{\mathcal{X}}
\def\mTheta{\mathbf{\Theta}}
\def\mL{\mathbf{L}}
\def\mQ{\mathbf{Q}}
\def\mA{\mathbf{A}}
\def\mD{\mathbf{D}}
\def\mLambda{\mathbf{\Lambda}}

\begin{document}
\maketitle

\begin{abstract}
Traditional methods for demand forecasting only focus on modeling the temporal dependency. However, forecasting on spatio-temporal data requires modeling of complex non-linear relational and spatial dependencies.
In addition, dynamic contextual information can have a significant impact on the demand values, and therefore needs to be captured. For example, in a bike-sharing system, bike usage can be impacted by weather.
Existing methods assume the contextual impact is fixed. However, we note that the contextual impact evolves over time. We propose a novel context integrated relational model, \emph{Context Integrated Graph Neural Network} (CIGNN), which leverages the temporal, relational, spatial, and dynamic contextual dependencies for multi-step ahead demand forecasting.
Our approach considers the demand network over various geographical locations and represents the network as a graph. We define a demand graph, where nodes represent demand time-series, and context graphs (one for each type of context), where nodes represent contextual time-series. 
Assuming that various contexts evolve and have a dynamic impact on the fluctuation of demand, our proposed CIGNN model employs a fusion mechanism that jointly learns from all available types of contextual information.
To the best of our knowledge, this is the first approach that integrates dynamic contexts with graph neural networks for spatio-temporal demand forecasting, thereby increasing prediction accuracy.
We present empirical results on two real-world datasets, demonstrating that CIGNN consistently outperforms state-of-the-art baselines, in both periodic and irregular time-series networks.
\end{abstract}

\begin{table*}[ht]

    \centering
    \begin{tabularx}{0.85\linewidth}{r ccccc cc} 
    
    \toprule
        & ARIMA  & VAR    & LSTM   & STGCN  & DCRNN  & WaveNet& \textbf{CIGNN} \\
    \midrule
\textsc{Temporal}     & \checkmark & \checkmark & \checkmark & \checkmark & \checkmark & \checkmark & \checkmark \\
\textsc{Spatial}      &            &            &            & \checkmark & \checkmark & \checkmark & \checkmark \\
\textsc{Relational}   &            &            &            & \checkmark & \checkmark & \checkmark & \checkmark \\
\textsc{Dynamic Contextual}   &            &            &            &            &            &            & \checkmark \\

    \bottomrule
    \end{tabularx}
    
    \caption{
    Qualitative comparison of CIGNN to previous methods.
    Notably, CIGNN is the only approach that learns and incorporates temporal, relational, spatial, and contextual dependencies.
    \label{tab:qualitative_cmp}}
\end{table*}

\section{Introduction}
Demand and supply forecasting with spatio-temporal data is widely studied in several areas including intelligent transportation systems~\cite{laptev2017time,2014stITS,wang2001spatio}
and infrastructure construction planning~\cite{csiszar2019urban,deb2017review}. For example, forecasting cellular data~\cite{2015basestation} demand in various places is critical for determining the ideal base station locations.
Demand and supply forecasting is typically formulated as a time-series prediction problem~\cite{STACM2011,willis1983spatial,ziat2017spatio}.
In spatio-temporal data analysis, there are multiple time-series recorded in various locations.
The dependence among them introduces the challenge of modeling relational and spatial dependencies for accurate forecasting.
Another challenge is to incorporate influence from factors that cause fluctuations in demand and supply.
Existing methods~\cite{ST-MGCN,DMVST-aaai2018} make a simplifying assumption that environmental contexts introduce a fixed impact.
However, we relax this assumption and consider dynamic contexts that have a time-evolving impact on demand and supply values. 

To address the aforementioned challenges,
we propose a novel GNN model, Context Integrated Graph Neural Network (\tb{CIGNN}), which learns and incorporates temporal, relational, spatial, and dynamic contextual dependencies for time-series forecasting.
Our approach represents the demand/supply network as a graph, and represents each type of context as a separate graph. Our method jointly learns a model to predict demand/supply and contextual time-series simultaneously. We design a fusion mechanism to model the contextual dependencies. To the best of our knowledge, this is the first work that integrates and exploits dynamic contexts in a unified way for spatio-temporal time-series predictions. We validate our model on two forecasting problems using real-world data: forecasting demand in a mobile call network and forecasting supply in a bike-sharing system on the \ti{CallMi} and \ti{BikeBay} datasets, respectively.
The main contributions are summarized as follows:
\begin{itemize}
    \item \tb{Modeling Temporal, Relational, Spatial, and Dynamic Contextual Dependencies}: 
    CIGNN performs demand forecasting by considering temporal, relational, spatial, and \emph{dynamic} contextual dependencies.
    Existing methods do not capture the dynamic context.
    Table~\ref{tab:qualitative_cmp} qualitatively compares the abilities of various methods to model these dependencies.
    
    \item \tb{Multi-source context Learning}: 
    Contrary to existing work that uses only a single contextual type, CIGNN is capable to integrate multiple types of contextual features to improve forecasting.

    \item \tb{Effectiveness}:
    CIGNN consistently outperforms previous methods with respect to both absolute error (MAE) and root mean square error (RMSE) for both \ti{CallMi} and \ti{BikeBay} datasets.
    CIGNN obtains average improvements of $5.7\%$ (MAE) and $9.4\%$ (RMSE) for \ti{CallMi}; and $4.4\%$ (MAE) and $2.3\%$ (RMSE) for \ti{BikeBay}.

\end{itemize}

\section{Related Work}
Traditional time-series prediction methods
such as Auto-Regressive Integrated Moving Average (ARIMA) and its variants~\cite{HamltontimeSeries}, suffer from several limitations. These methods 
(1) do not capture relational and spatial correlations, 
(2) cannot handle time-series with irregularities, 
and (3) generally perform poorly on long-term forecasting.

Forecasting based on deep learning
allows the integration of complex temporal, relational, spatial, and contextual correlations to infer predictions.
Previous work in spatio-temporal study leverages Convolutional Neural Networks (CNN)~\cite{DMVST-aaai2018,miao2019st} and Graph Neural Networks (GNN)~\cite{li2015gated,GNNSurvey2009,wu2020comprehensive,bruna2013spectral} to capture the spatial dependency.
Recurrent Neural Networks (RNN) and their variants~\cite{seq2seqNIPS2014,wu2017uapd} have also been used to capture temporal correlations.

\Paragraph{Graph Neural Networks}
have shown unprecedented performance in different domains of study
~\cite{spatial_temporal_attention,rossi2018ker,bruna2013spectral,kipf2017gcn}.
For instance,
GraphSAGE~\cite{hamilton2017inductive} and FastGCN~\cite{chen2018fastgcn}, sample and aggregate neighborhood information to perform a 
graph classification task.
Another line of study leverages GNN to perform forecasting~\cite{zheng2020gman,zhang2020spatio,song2020spatial,chen2019multi}.
Most of these works targets traffic forecasting improvements.
For example, a diffusion convolution strategy has been proposed to predict the speed at one location by considering the speeds in proximity~\cite{li2018dcrnn_traffic}.
Another work utilized a GNN on both crime and traffic forecasting.~\cite{Wang2018GraphBasedDM} 
Other applications include
city-wide bike demand forecasting~\cite{defferrard2016convolutional,lin2018predicting} and weather prediction~\cite{wilson2018low}.
However, these methods ignore contextual information, thereby limiting their predictive performance.

\Paragraph{Contextual features} have a significant impact on demand and supply.
For instance, ride-hailing demand is highly sensitive to precipitation.
\begin{figure*}
    \centering
    \includegraphics[width=\linewidth]{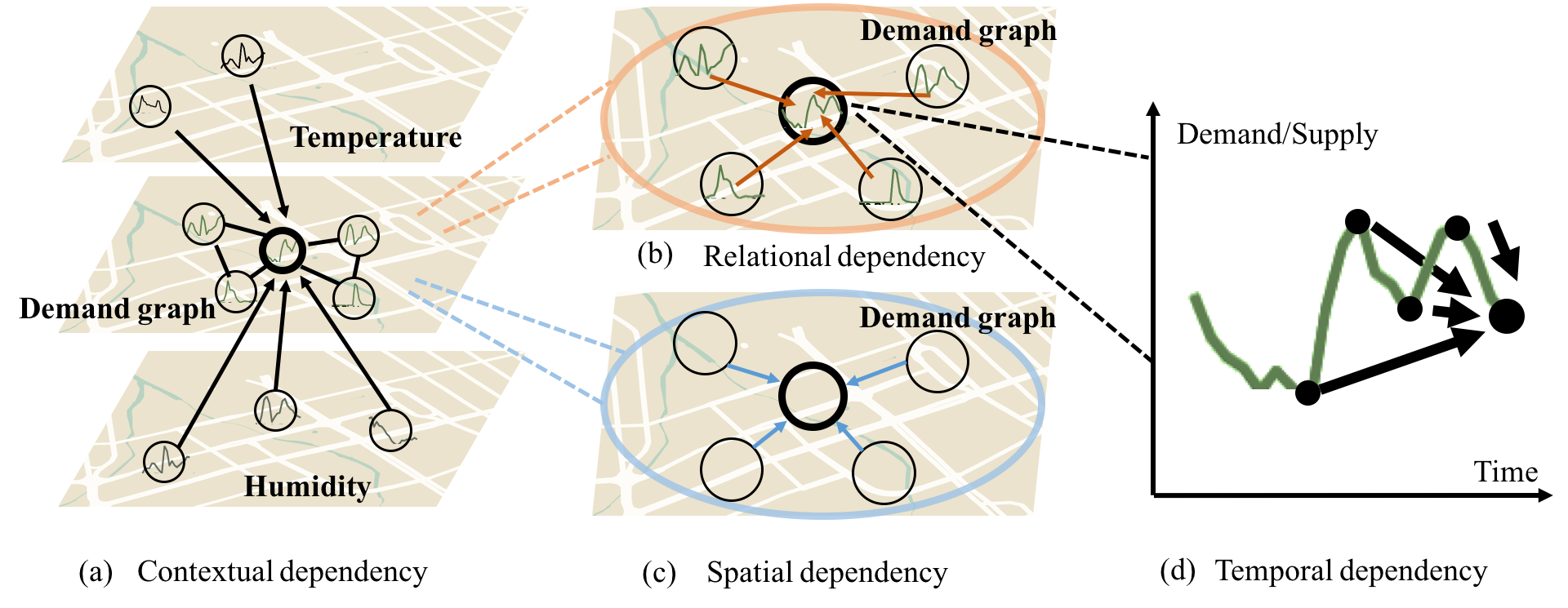}
    \caption{An illustration of spatial, temporal, relational, and contextual dependencies using only a node for simplicity.
    (a) Contextual dependency represents the impact of context (humidity and temperature) graphs on the demand graph.
    (b) Relational dependency represents the time-series correlation between nodes within
    a graph.
    (c) Spatial dependency represents the spatial correlation between nodes within
    a graph.
    (d) Temporal dependency represents the influence from history on future values.}
    \label{fig:intro_dependencies}
\end{figure*}
Existing approaches~\cite{ParkingAvailability} assume a \emph{static context},
which leads to a loss of 
information.
By contrast, our approach integrates dynamic information to capture contextual impact.
Moreover, existing work~\cite{DMVST-aaai2018} incurs an increased cost of feature selection since their features are manually designed and integrated into the model.
Our model, however, can handle various types of contextual information from multiple contextual sources simultaneously.

\section{Context Integrated Graph Neural Network (CIGNN)}
\label{sec:cignn}
We propose a novel context integrated graph model, which leverages the temporal, relational, spatial, and contextual dependencies for demand and supply forecasting.
Our method represents the demand network over locations as a graph. We define a demand graph, where nodes represent demand time-series, and context graphs (one graph for each type of context), where nodes represent contextual time-series. 

We introduce our model \emph{Context Integrated Graph Neural Network} (CIGNN) using a bike-sharing system as a running example. Given previous bike supply observations at various stations, CIGNN predicts the future supply by considering the following dependencies (also illustrated in \figref{intro_dependencies}):

\begin{definition}[Temporal dependency]
\label{def:temporal}
Given a time-series $\mb{x}=[x_1  x_2  \cdots  x_T]$ where $T$ is the length of the time-series.
The temporal dependency is a mapping $f$ from some observations prior to $x_t$ (i.e. $w$ = time-series lag) to $x_t$:
\begin{align}
    [x_{t-1} x_{t-2} \cdots x_{t-w}] \xrightarrow[]{f(\cdot)} x_t
    \label{eq:temporal}
\end{align}
\end{definition}
The temporal dependency implies that future bike supply in a location relies on its historical supply.

The supply at a location also relies on those at others.
We hereby define the relational and spatial dependency:
\begin{definition}[Relational dependency]
\label{def:relationaldep}
Given a graph $G=(V, E)$, where $V$ denotes the node set and $E$ the edge set.
Each node represents a bike station associated with a time-series that indicates the supply.
Then, we define edges representing relational dependencies:
\begin{align}
    \!E=\{(i, j)\mid \forall (i, j) \in \card{V} \!\times \!\card{V}, \ti{s.t.}\> \mb{\mK}(\mb{x}_i, \mb{x}_j) \!\geq \!\lambda_r\}
    \label{eq:relational}
\end{align}
where $\mb{x}_i, \mb{x}_j$ denote the time-series associated to nodes $i$ and $j$, respectively. $\card{V}$ is the number of nodes and $\mK$ denotes a metric that measures the correlation between series.
\end{definition}
Edges with a large weight $\mb{\mK}(\mb{x}_i, \mb{x}_j)$ imply dependency when the weight is greater than or equal to a threshold $\lambda_r$.

The key idea behind relational dependency is that similarly behaving time-series tend to be correlated in the future,
For spatial dependency, we construct a graph differently.
\begin{definition}[Spatial dependency]
\label{def:spatialdep}
Given a graph $G=(V, E, \mA)$
where $V$ denotes the node set and $E$ the edge set.
Each node represents a location of a bike station. The adjacency matrix $\mA$ denotes the distance between nodes.
We define the edges to represent spatial dependency:
\begin{align}
    E=\{(i, j)\mid \forall (i, j) \in \card{V} \times \card{V}, \ti{s.t.}\> \mA_{ij} \geq \lambda_s\}
    \label{eq:spatial}
\end{align}
\end{definition}
Edges with a large weight $\mA_{ij}$ indicate potential dependency when the weight is greater than or equal to a threshold $\lambda_s$.

\begin{definition}[Contextual dependency]
\label{def:contextualdep}
Let a supply graph $G_s=(V_s, E_s)$ with a set of nodes $V_s$ representing the bike stations that connected by edges in $E_s$.
In addition, let a context graph $G_c=(V_c, E_c)$ with a set of nodes $V_c$ representing the locations where contextual features are recorded (\eg weather stations recording humidity), and connected by edges in $E_c$. Note that $E_s$ and $E_c$ are edges derived either from \defref{relationaldep} or \defref{spatialdep}. We denote $E_{sc}$ as edges connecting nodes in $V_s$ with nodes in $V_c$.
\begin{align}
    E_{sc}=\{(i, j)\mid \forall (i, j) \in \card{V_s} \times \card{V_c}\}
    \label{eq:contextual}
\end{align}
To account for possibly more than one contextual type (\eg if there are n contextual types), the definition can be extended to include all contextual types: $V_{c}=V_{c1}
\cup\ldots\cup V_{cn}$,
$E_{c}=E_{c1}
\cup\ldots\cup E_{cn}$, for contextual types $c1,\ldots,cn$.
\end{definition}

\begin{table}[ht]
\begin{center}
    \centering
    {
    \begin{tabular}{rl}
    \toprule
        \tb{Symbol} & \tb{Description} \\
    \midrule
    $\mathcal{G}, M=|\mathcal{G}|$ & a set of graphs and the number of graphs                             \\
    $T, T_{w}, T_{h}$ & length of time-series, window and horizon                                       \\
    \midrule
    $G_i$ & the $i^{th}$ graph, where $i \in \{1, \cdots, M\}$                                            \\
    ${V}_i, {E}_i, \mb{A}_i$ & nodes, edges and matrix in $i^{th}$ graph                    \\
    $N_i=|{V}_i|$, $P_i$ & number of nodes and features in $i^{th}$ graph                       \\
    $\mathcal{X}_i$ 
    & the graph signal of the $i^{th}$ graph \\
    \midrule
    $G, \mathcal{X}$ & a graph and its graph signal                                                       \\
    $\mb{A, D}$ & Adjacency matrix and degree matrix                                     \\
    $\mb{L, I}$ & Laplacian matrix and the identity matrix\\
    \midrule
    $\mb{\Theta}_{*G}$ & the graph convolution layer                                                \\
    $\mb{\Phi}$ & proposed fusion layer                                                               \\
    $\theta,\mb{W, b, z}$ & trainable parameters\\
    \bottomrule
    \end{tabular}}
\end{center}
\label{tab:notation}
\caption{Summary of notation}
\end{table}
In a bike-sharing system, the bike supply is impacted by weather such as temperature and precipitation, which are recorded in weather stations.
We build a graph for temperature time-series and one for precipitation time-series.

\begin{figure*}[ht]
    \centering
    \includegraphics[width=\linewidth]{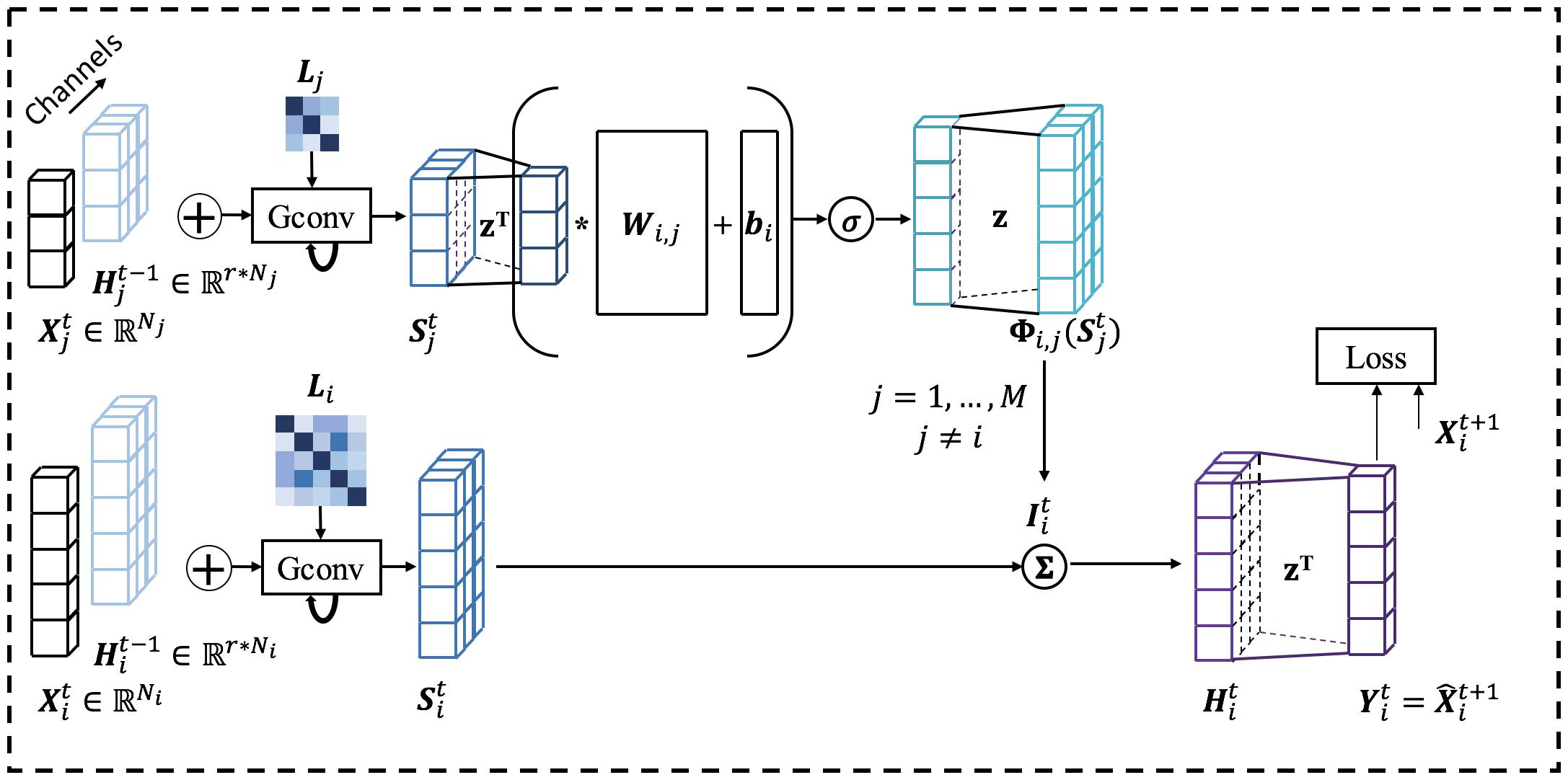}
    \hspace*{\fill}
    \caption{
    An overview of our CIGNN unit.
    Due to space limitations, only two graphs are displayed in the figure. In practice, CIGNN learns $\mb{\Phi}$ for every combination of $(i, j)$ where $ i, j \in {1, 2, ..., M}, i\neq j$. }
    \label{fig:architecture}
\end{figure*}

\subsection{Problem Formulation}
\label{subsec:formulation}
Demand and supply forecasting in spatio-temporal data is a time-series prediction task where time-series are associated with locations.
The task aims to learn a function $f$ that predicts future values for each time-series.
We refer to time-series in a graph as a \ti{graph signal}~\cite{furutani2019graph}. A graph signal consists of signals from the same (graph) type.
The demand/supply network and the $M-1$ context networks are represented as a set of $M$ graphs: $\mathcal{G}=\{G_1, G_2, \dots, G_M\}$.
Each graph either represents the demand/supply or a contextual type.
For instance, to forecast the bike supply, the derived graphs include a bike supply graph, a temperature graph, and a humidity graph.
The $i^{th}$ graph is denoted as $G_i=({V}_i, {E}_i, \mA_i)$, where ${V}_i$ and ${E}_i$ represent the set of nodes and the set of edges in $G_i$, respectively.
Nodes represent time-series with associated locations and edges represent the dependency between nodes.
$\mA_i$ is the adjacency matrix that encodes geographical distance.

The graph $G_i$ is associated with its corresponding graph signal $\mX_i \in \mR^{T \times N_i \times P_i}$,
where $T$ denotes the time span of interest, $N_i= |{V}_i|$ is the number of nodes, $P_i$ is the number of node features.
We use a subscript to denote which graph is referred, and a superscript to denote the time. For example, $\mX_i^t \in \mR^{N_i\times P_i}$ is the graph signal $G_i$ at $t$.

Given graphs and graph signals, a multi-step ahead time-series prediction task regarding $T_{w}$ past observations and a horizon $T_{h}$ is formulated as:
\begin{align*}
\left[ \mX_i^{t-T_{w}+1}\mX_i^{t-T_{w}+2} \dots \mX_i^t; G_i \right] \xrightarrow[]{f(\cdot)} \left[ \mX_i^{t+1}\mX_i^{t+2}\dots \mX_i^{t+T_h}\right]
\\
i \in [1, \dots, M]
\end{align*}

\subsection{Graph Convolution to Exploit Relational and Spatial Features}
\label{subsec:gcn}
Graph convolution is powerful to model the relational dependency and the spatial dependency~\cite{wu2020comprehensive}.
Given a graph $G$ (constructed with either~\defref{relationaldep} or~\defref{spatialdep}) and its graph signal $\mX$, the graph convolution layer is defined as:
\begin{align}
    \mTheta_{*G}\mX &= \mTheta(\mL)\mX \nonumber \\
    &= \mTheta(\mb{Q\Lambda}\mQ^T)\mX  \nonumber \\
    &= \mQ\mTheta(\mLambda)\mQ^T \mX
    \label{eq:eigendecompose}
\end{align}
where $\mQ$ is composed of eigenvectors of the graph Fourier transform. $\mLambda$ is a diagonal matrix where each element is an eigenvalue of the normalized Laplacian matrix $\mL$, defined with adjacency matrix $\mA$ and diagonal degree matrix $\mD$:
\begin{equation}
    \mL = \mD^{-\frac{1}{2}} (\mD - \mA) \mD^{-\frac{1}{2}}
\end{equation}

\Paragraph{Strengthen Locality with Chebyshev Polynomials Approximation}. Assuming that neighbor nodes have higher impacts, we reinforce the weights from the local neighborhood. The graph kernel $\mTheta(\mLambda)$ is extended to a series of polynomial bases as:
\begin{equation}
    \mTheta(\mLambda) = \sum_{k=0}^{K-1}\theta_k\mLambda^k
    \label{eq:GFourier}
\end{equation}
where $K$ regulates the locality radius of nodes, and $\mathbf{\theta}\in \mR^k$ is a vector parameter of polynomial coefficients.
A truncated Chebyshev polynomial expansion is adapted with Eq.~\ref{eq:GFourier} for computation efficiency:
\begin{align}
    \mTheta_{*G}\mX &=\mTheta(\mL)\approx\sum_{k=0}^{K-1}\theta_k T_k(\mb{\tilde{L}})\mX \\ 
    \mb{\tilde{L}} &= \frac{2\mL}{\lambda_{max}} - \mb{I} \label{eq:scaledL}
     = \mL - \mb{I}, \text{assuming}\enspace\lambda_{max}\!=\!2
\end{align}\noindent
where $T_k(\mb{\tilde{L}})$ is the $k^{th}$ Chebyshev polynomial at the scaled Laplacian $\mb{\tilde{L}}$ and $\mb{I}$ is the identity matrix~\cite{defferrard2016convolutional,tang2019chebnet}.
$\lambda_{max}$ is the greatest eigenvalue of $\mL$, which is assumed as 2 to simplify the Eq.~\ref{eq:scaledL}, as parameters can adapt to the change in scale during training~\cite{kipf2017gcn}.
The Chebyshev polynomial reduces the time complexity from $\mathcal{O}(N^3)$ to $\mathcal{O}(K|\mathcal{E}|)$
where $\mathcal{E}$ is the number of non-zero edges. 
                     
Graph convolution is separately applied on each graph:
\begin{equation}
    \mTheta_{*G_i}(\mX_i) = \sum\limits_{k=0}^{K-1}\theta_{k, i}\mL_i^ k\mX_i, \quad i=1, 2, \dots, M
    \label{graph-c-operation}
\end{equation}

\subsection{
Network Architecture}
\label{subsec:CIGNN}
CIGNN is composed of units that take inputs at the current time step $\mb{X}_i^t \in \mR^{N_i \times P_i}$ and hidden states from previous time step $\mb{H}_i^{t-1} \in \mR^{r \times N_i \times P_i}$.
With inputs and hidden states, CIGNN give hidden states of current step as $ \mb{H}_i^{t-1} \in \mR^{r \times N_i \times P_i}$, where $r$ is the number of neurons.
CIGNN is formulated as follows:
\begin{align}
    \mb{r}_i^t &= \sigma(\mb{FC}_r(\mTheta_{r*G_i}[\mb{X}_i^t\oplus\mb{H}_i^{t-1}])) \label{eq:gru1} \\
    \mb{u}_i^t &= \sigma(\mb{FC}_u(\mTheta_{u*G_i}[\mb{X}_i^t\oplus\mb{H}_i^{t-1}])) \label{eq:gru2} \\
    \mb{C}_i^t &= tanh(\mb{FC}_C(\mTheta_{C*G_i}[\mb{X}_i^t\oplus(\mb{r}_i^t\odot\mb{H}_i^{t-1})])) \label{eq:gru3} \\
    \mb{S}_i^t &= \mb{u}_i^t\odot\mb{H}_i^{t-1} + (1-\mb{u}_i^t)\odot\mb{C}_i^t \label{eq:gru4} \\
    \mb{I}_i^t &= \sum_{\substack{j=1 \> j\neq i}}^M\mb{\Phi}_{i,j}(\mb{S}_j^t) \label{eq:gruI}\\
    \mb{H}_i^t &= \mb{S}_i^t + \mb{I}_i^t
\end{align}
where Eq.~\ref{eq:gru1}-Eq.~\ref{eq:gru4} are similar to the structure of Gated Recurrent Units
~\cite{GRUchung2014}. $\mb{r}$ and $\mb{u}$ denote the reset gate and update gate.
$\mb{FC}$ is a dense layer. $\mTheta$ is the graph convolution layer,
which captures either relational dependency or spatial dependency determined by what graph is used.
The learned state $\mb{S}$ incorporates both temporal, relational or spatial dependency.
$\mb{\Phi}$ is our proposed fusion layer that captures the contextual dependency across graphs, as defined in~\defref{contextualdep}.
The subscript denotes different sets of parameters. 
The operator $\oplus$ and $\odot$ denotes concatenation operation and element-wise multiplication, respectively.

The fusion layer incorporate a graph with impacts from contextual graphs. 
$\mb{I}_i$ in Eq.~\ref{eq:gruI} captures impacts from other graphs on the $i^{th}$ graph, as shown in~\figref{architecture}.
The structure of the fusion layer is designed in an interleaved manner:
\begin{equation}
    \mb{\Phi}_{(i, j)}(\mb{S}_j) = \sigma(\mb{z}[(\mb{W}_{i,j}^T \mb{z}^T \mb{S}_j^t + \mb{b}_j])
\end{equation}
\noindent
where $\sigma$ denotes the sigmoid function and $\mb{S}_j$ represents the hidden state 
from graph $j$. 
The weight parameters $\mb{W} \in \mR^{N_j \times P_j \times P_i \times N_i}, \mb{b} \in \mR^{N_i \times P_i}$ 
extracts relations between time-series across graphs, \ie the contextual dependency. Note that $W$ has a similar form as $E_{sc}$ defined in Eq.~\ref{eq:contextual} when we consider the demand/supply graph and its contextual graphs.
The parameter $\mb{z} \in \mR^{r}$ is a mapping vector.
Given the hidden state $\mb{H}_i^t$, the forecasting is conducted as:
\begin{equation}
\hat{\mb{X}}_i^{t+h} = \mb{z}_h^T \mb{H}_i^{t}, \> h=1, 2, \dots, T_h
\label{eq:final}
\end{equation}\noindent
where there is a $\mb{z_h}$ for each horizon.
Notice that the Eq.~\ref{eq:final} models temporal dependency as in~\defref{temporal}.
A goal function is then designed to minimize the prediction errors:
\begin{align}
    \argmin_{
        \theta_{r}, \theta_{u}, \theta_{C},
        \mb{W}, \mb{b}, \mb{z}
        }\left|\hat{\mb{X}}_i^{t+h}-\mb{X}_i^{t+h}\right|,  \\\nonumber
        \> i=1, 2, \dots, M,
        \qquad\! h=1, 2, \dots, T_h
\end{align}

\begin{figure}[ht]
    \centering
    \includegraphics[width=\linewidth]{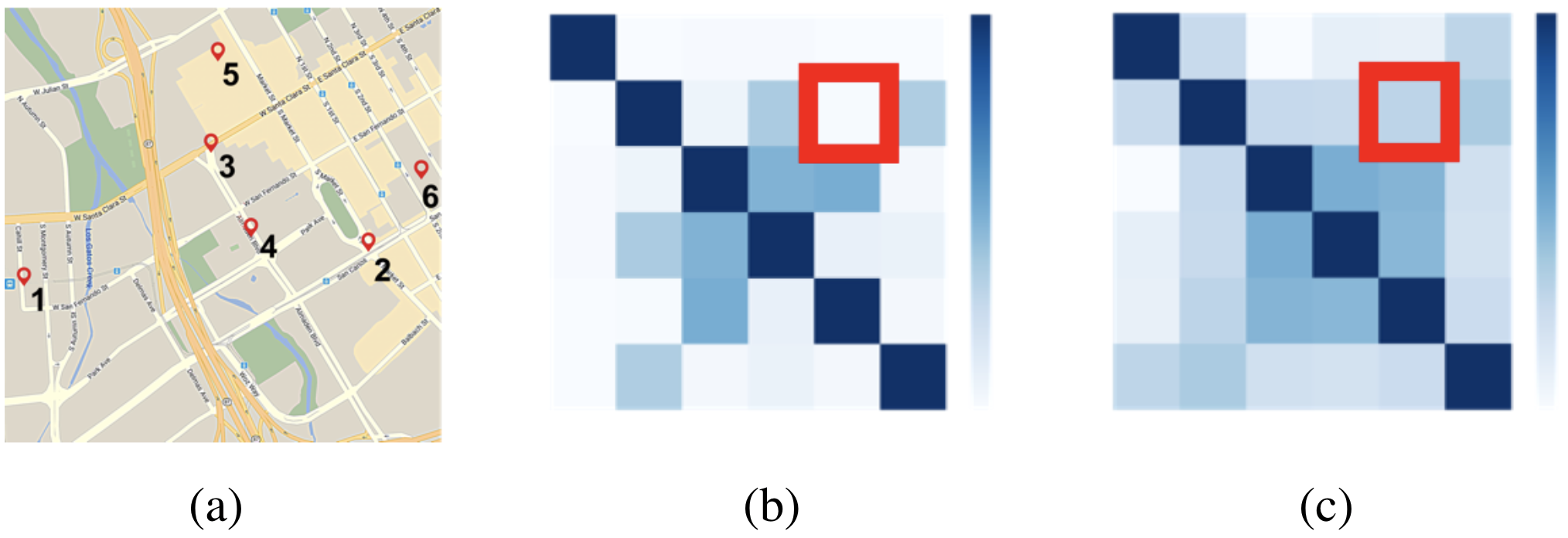}
    \caption{A comparison between Gaussian kernel matrix and Relational matrix. Dark color indicates higher correlations. 
    (a). Some bike docks in San Jose.
    (b). The Gaussian kernel matrix. 
    (c). The relational correlation matrix. The value between station 2 and 5 (marked by the red squares) is higher. 
    }
    \label{fig:adj_example_comp}
\end{figure}

\section{Experiments}
\label{sec:exp}
In this section, we first introduce how the graphs are constructed, then we describe our experimental setup including two real-world datasets and previous approaches.
Finally, we discuss our experimental results.
\subsection{Graph Construction}
We introduce two existing ways to construct a graph. One uses geographical distance to model spatial dependency.
the other models the implicit correlations between time-series. Unless otherwise stated, we denote the adjacency matrix as $\mA \in \mR^{N \times N}$, and $\mA_{ij}$ is an element.

\Paragraph{Distance-based Gaussian Kernel Matrix}. Most existing work~\cite{li2018dcrnn_traffic,STGCN-ijcai2018} derive an adjacency matrix with a truncated Gaussian kernel, such that $\mA_{i,j}$ have higher values if node $i$ and node $j$ are nearer:
\begin{equation}
\mA_{ij} =  
\begin{cases}
    exp({-\frac{d(i, j)^2}{\sigma^2}}),&\text{if} \> d(i, j) \leq \kappa\\
    0,              & \text{otherwise}
\end{cases}
\end{equation}
where $d(i,j)$ denotes the geographical 
$\sigma$ denotes the standard deviation of distances
and $\kappa$ is a threshold.

\Paragraph{Relational Matrix based on Correlation Coefficients}.
We observe in empirical studies that the Gaussian kernel matrix can fail to capture the hidden relational correlations between distant nodes. 
For example, in \figref{adj_example_comp}a, station 2 and station 5 are far from each other and the Gaussian Kernel matrix (\figref{adj_example_comp}b) suggests they have low correlations.
However, they are highly correlated (\figref{adj_example_comp}c) in reality since people frequently commute between them along the straight road. Thus, we use the Detrended Cross-Correlation Analysis coefficient (DCCA coefficient)~\cite{kristoufek2014measuring,ide2017detrended} to construct the \ti{relational correlation matrix}.
    
The DCCA coefficient~\cite{xu2010modeling} is a metric that infers correlations between series. It combines detrended cross-correlation analysis (DCCA) and detrended fluctuation analysis (DFA).
Given two time-series $\mb{x}, \mb{y} \in \mR^T$ and a window length $l$, the DCCA coefficient is defined as:
\begin{equation}
    \rho_{DCCA}(\mb{x}, \mb{y}, l) = \frac{F_{DCCA}^2(\mb{x},\mb{y},l)}{F_{DFA}(\mb{x},l)F_{DFA}(\mb{y},l)}
\end{equation}
where numerators and denominators are the average covariances and variances of the $T-s+1$ windows (partial sums):
\begin{align}
    F_{DCCA}^2(\mb{x}, \mb{y}, l) &= \frac{\sum_{s=1}^{T-l+1} f_{DCCA}^2 (\mb{x}, \mb{y}, s)}{T-l}\\
    F_{DFA}^2(\mb{x}, l) &= \frac{\sum_{s=1}^{T-l+1} f_{DFA}^2 (\mb{x}, s)}{T-l}
\end{align}
The partial sums are calculated with sliding windows across $\mb{x}$ and $\mb{y}$. For each window with starting index $s$:
\begin{align}
    \label{eq:dccawindow}
    f_{DCCA}^2(\mb{x}, \mb{y}, s) &= \frac{\sum_{t=s}^{s+l-1}(\mb{x}^t - \bar{\mb{x}}_s)(\mb{y}^t - \bar{\mb{y}}_s)}{l-1} \\
    f_{DFA}^2(\mb{x}, s) &= \frac{\sum_{t=s}^{s+l-1}(\mb{x}^t - \bar{\mb{x}}_s)^2}{l-1}
\end{align}
where $\bar{\mb{x}}_s$ is the average value of the window started with $s$.

The matrix is constructed with the pairwise correlations:
\begin{equation}
\mA_{ij} =  
\begin{cases}
    \rho_{DCCA}(\mb{x}, \mb{y}, l),&\text{if} \> \rho_{DCCA}(\mb{x}, \mb{y}, l) \geq 0\\
    0,              & \text{otherwise}
\end{cases}
\end{equation}

\begin{table*}[bt]
    \centering
    {
    \begin{tabular}{c c rrrrrrrr}
    \toprule
    \multicolumn{10}{c}{\tb{\ti{CallMi}}} \\
    Horizon             & Metrics & HA    & ARIMA & VAR   & LSTM  & STGCN & DCRNN & WaveNet & \tb{CIGNN} \\
    \midrule
    \multirow{2}{*}{1}  & MAE     & 17.15 & 14.42 & 18.54 & 13.51 & 11.35 & 10.41 & 9.48  & \tb{8.89}  \\
                        & RMSE    & 38.80 & 24.26 & 31.30 & 25.04 & 20.46 & 19.44 & 18.18 & \tb{16.82} \\
    \midrule
    \multirow{2}{*}{2}  & MAE     & --    & 26.78 & 27.01 & 17.05 & 20.48 & 16.59 & 12.72 & \tb{11.72} \\
                        & RMSE    & --    & 44.27 & 47.14 & 30.10 & 35.63 & 33.59 & 26.23 & \tb{22.84} \\
    \midrule
    \multirow{2}{*}{3}  & MAE     & --    & 38.12 & 34.49 & 19.02 & 33.14 & 22.60 & 15.38 & \tb{14.86} \\
                        & RMSE    & --    & 61.43 & 60.13 & 35.04 & 40.01 & 55.03 & 32.09 & \tb{29.59} \\

    \midrule
    \midrule
    \multicolumn{10}{c}{\tb{\ti{BikeBay}}} \\
    Horizon             & Metrics & HA    & ARIMA & VAR   & LSTM  & STGCN & DCRNN & WaveNet & \tb{CIGNN} \\
    \midrule
    \multirow{2}{*}{1}  & MAE     & 22.70 & 7.62  & 8.04  & 19.91 & 6.80  & 6.55  & 7.00    & \tb{6.37}  \\
                        & RMSE    & 28.92 & 13.35 & 18.72 & 25.34 & 11.98 & 12.37 & 13.33   & \tb{11.75} \\
    \midrule
    \multirow{2}{*}{2}  & MAE     & --    & 11.70 & 12.06 & 20.83 & 10.76 & 10.13 & 11.01   & \tb{9.68}  \\
                        & RMSE    & --    & 18.62 & 29.42 & 26.67 & 16.98 & 17.65 & 19.56   & \tb{16.60} \\
    \midrule
    \multirow{2}{*}{3}  & MAE     & --    & 14.12 & 14.34 & 21.29 & 13.41 & 12.56 & 13.69   & \tb{11.90} \\
                        & RMSE    & --    & 21.19 & 35.05 & 27.73 & 19.71 & 20.52 & 22.25   & \tb{19.18} \\
    \bottomrule
    \end{tabular}}
\caption{A comparison using 6 observed data points to predict 3 steps ahead. MAE and RMSE of three horizons, their average, and average error reduction over VAR as a baseline for \ti{CallMi} and \ti{BikeBay}
\label{tab:results_CallMi_BikeBay}}
\end{table*}

\subsection{Experimental Setup}
To verify the effectiveness of CIGNN, we conduct experiments with two public real-world datasets.
For both datasets, the following sets of hyperparameters are used: 0.01 (learning rate), 32 (number of neurons), 0.1 (learning rate decay ratio for every 10 epochs).
We train for a maximum of 100 epochs using the Adam optimizer~\cite{kingma2014adam} and adapt an early stop strategy if the validation loss does not decrease for 10 consecutive epochs. 
All experiments are implemented using Python Tensorflow (v1.14)
and run on Ubuntu 16.04 with 8 CPU cores and a memory of 32G.
\begin{itemize}
    \item \tb{Mobile Call Demand in Milan \ti{(CallMi)}} \cite{milanCallData}: \ti{CallMi} contains call demand data in Milan from Nov. 2013 to Dec. 2013.
    The dataset 
    contains temperature and humidity as contextual features. 
    The city is partitioned into grids in the raw dataset, however, some grids have very few records.
    Therefore, we cluster grids into $162$ mobile call nodes. There are $5$ temperature nodes and $4$ humidity nodes. Each of contextual nodes corresponds to a weather station. The time interval is an hour.
    \item \tb{Bike-sharing Supply in the Bay Area \ti{(BikeBay)}} \cite{bikingData}: \ti{BikeBay} contains the bike supply data in $70$ dock stations in the Bay Area. The dataset was recorded from Aug. 2013 to Aug. 2015. It contains weather conditions as contextual data. There are $3$ nodes for temperature, humidity, dew, sea level, and wind speed, respectively. The time interval is two hours.
\end{itemize}
\begin{figure}[t]
    \centering    \includegraphics[width=\linewidth]{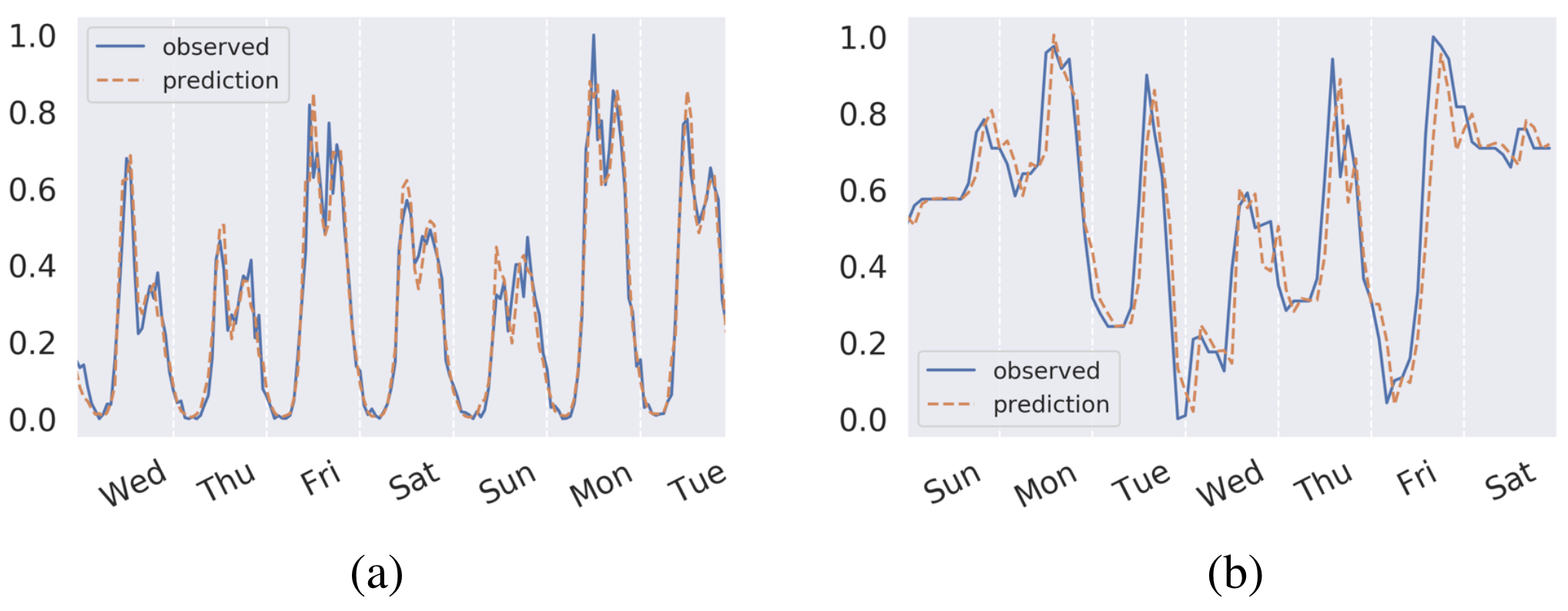}
    \caption{A comparison of time-series pattern. (a) The call demand during Dec. 25-31. (b) The bike supply in San Francisco during Aug. 25-31. Values are normalized for display. Note that \ti{CallMi} shows more periodicity.}
    \label{fig:ts_periodicity_comp}
\end{figure}
\Paragraph{Diversity of Datasets regarding Periodicity.}
The call demand time-series exhibit periodicity
(\figref{ts_periodicity_comp}a), while the bike supply time-series shows more irregularities (\figref{ts_periodicity_comp}b).
We chose datasets to demonstrate CIGNN's capability to predict well on time-series with or without periodicity.
The data are split chronologically into training, validation and testing sets in a ratio of 70\%:10\%:20\%.
When using the relational matrix, the window length $l$ in Eq.~\ref{eq:dccawindow} is set as 4.
The graph convolution step in Eq.~\ref{graph-c-operation} is set as $1$, for a previous study~\cite{RyanGCN} shows that a small step of graph convolution is more effective. 
Mean Absolute Error (MAE) is used as a loss measure to update parameters for all horizons in the training set.
In evaluation, both MAE and Root Mean Square Error(RMSE) are calculated and compared:
\begin{align*}
    \ti{MAE}&=\frac{1}{|\mb{\Omega}|}\sum_{t\in\mb{\Omega}}|\mX_i^t - \hat{\mX_i^t}|
    \\
    \ti{RMSE}&=\sqrt{\frac{1}{|\mb{\Omega}|}\sum_{t\in\mb{\Omega}}(\mX_i^t - \hat{\mX_i^t})^2},
    \quad i=1, 2, \dots, M
\end{align*}
where $\mb{\Omega}$ denotes the timestamps of measured samples.

\begin{table}[!b]
    \centering
    {    
    \begin{tabular}{c c rrrr}
    \toprule
    \multicolumn{6}{c}{\tb{\ti{BikeBay}}}                               \\
    H            & Metrics & STGCN & DCRNN & WaveNet & CIGNN      \\
    \midrule
    \multirow{2}{*}{1} & MAE     & 7.50  & 6.44  & 19.24   & \tb{6.38}  \\
                       & RMSE    & 12.37 & 11.99 & 26.75   & \tb{11.80} \\
    \midrule
    \multirow{2}{*}{2} & MAE     & 10.71 & 9.72  & 19.77   & \tb{9.60}  \\
                       & RMSE    & 16.63 & 16.72 & 27.38   & \tb{16.47} \\
    \midrule
    \multirow{2}{*}{3} & MAE     & 13.11 & 11.91 & 20.32   & \tb{11.75} \\
                       & RMSE    & 19.33 & 19.33 & 28.02   & \tb{19.01} \\
    \midrule
    \multirow{2}{*}{4} & MAE     & 16.62 & 13.67 & 20.68   & \tb{13.47} \\
                       & RMSE    & 22.77 & 21.22 & 28.36   & \tb{20.85} \\
    \midrule
    \multirow{2}{*}{5} & MAE     & 16.44 & 15.06 & 20.94   & \tb{14.83} \\
                       & RMSE    & 23.29 & 22.64 & 28.57   & \tb{22.24} \\
    \midrule
    \multirow{2}{*}{6} & MAE     & 17.56 & 16.12 & 21.12   & \tb{15.90} \\
                       & RMSE    & 24.83 & 23.67 & 28.68   & \tb{23.31} \\
    \bottomrule
    \end{tabular}}
\caption{A comparison of deep learning methods using 24 observed data points to predict 6 steps ahead on \ti{BikeBay}.
\label{tab:results_BikeBay_lag24_hor6}}
\end{table}

\subsection{Baseline Methods}
\label{subsec:baselines}
\noindent
We compare CIGNN to state-of-the-arts methods.
However, we did not compare to~\cite{DMVST-aaai2018} since it is only applicable on grid formatted data.
\begin{enumerate}
    \item \tb{HA} Historical Average. A prediction for a given time is the average of previous values at that same time (and day) over the past four weeks.
    \item \tb{ARIMA} Auto-Regressive Integrated Moving Average.
    The orders are $(3, 0, 1)$, as in~\cite{li2018dcrnn_traffic}.
    \item \tb{VAR} Vector Auto-Regressive is a multi-variate model that generalizes ARIMA to have multiple evolving variables.
    \item \tb{MM-LSTM} Multi-step multi-variate LSTM. The number of neurons is set as 64.
    \item \tb{DCRNN} \cite{li2018dcrnn_traffic}: Diffusion Convolution Recurrent Neural Network  models spatial correlations with a diffusion process.
    \item \tb{STGCN} \cite{STGCN-ijcai2018}: Spatio-Temporal Graph Convolutional Network
    models the spatial and temporal dependencies with a gating mechanism.
    \item \tb{Graph WaveNet} \cite{graphwavenet-ijcai2019} 
    Graph WaveNet leverages a self-adaptive adjacency matrix design to exploit spatial dependencies.
\end{enumerate}

\subsection{Results}
\Paragraph{Comparisons for multi-step ahead prediction} Following common practice~\cite{li2018dcrnn_traffic,STGCN-ijcai2018,graphwavenet-ijcai2019}, we
use six past observations to predict three steps ahead.
Results are shown in \tabref{results_CallMi_BikeBay}.
MAE and RMSE are evaluated for each horizon and on the average across horizons.
Using dataset \ti{BikeBay}, we further conduct more experiments with more data points (24) and a greater horizon (6) on the deep learning methods, as shown in \tabref{results_BikeBay_lag24_hor6}.
We observe the following:
\begin{itemize}
    \item Deep learning based methods outperform traditional methods (HA, ARIMA and VAR) due to the latter methods' limit of only modeling temporal dependency. Besides, HA can only predict one step ahead, and ARIMA fails to exploit the interactions across time-series.
    \item For deep learning methods, WaveNet outperforms DCRNN on CallMi but not on BikeBay, while CIGNN consistently outperforms STGCN, DCRNN and WaveNet. CIGNN also outperforms baselines when more observations and a larger horizon are considered.
    \item CIGNN outperforms previous best state-of-the-art models on each dataset (improves over WaveNet on CallMi by $5.7\%$ MAE and $9.4\%$ RMSE and over DCRNN on BikeBay by $4.4\%$ MAE and $2.3\%$ RMSE).
\end{itemize}

\begin{figure}[!t]
    \centering
    \includegraphics[width=\linewidth]{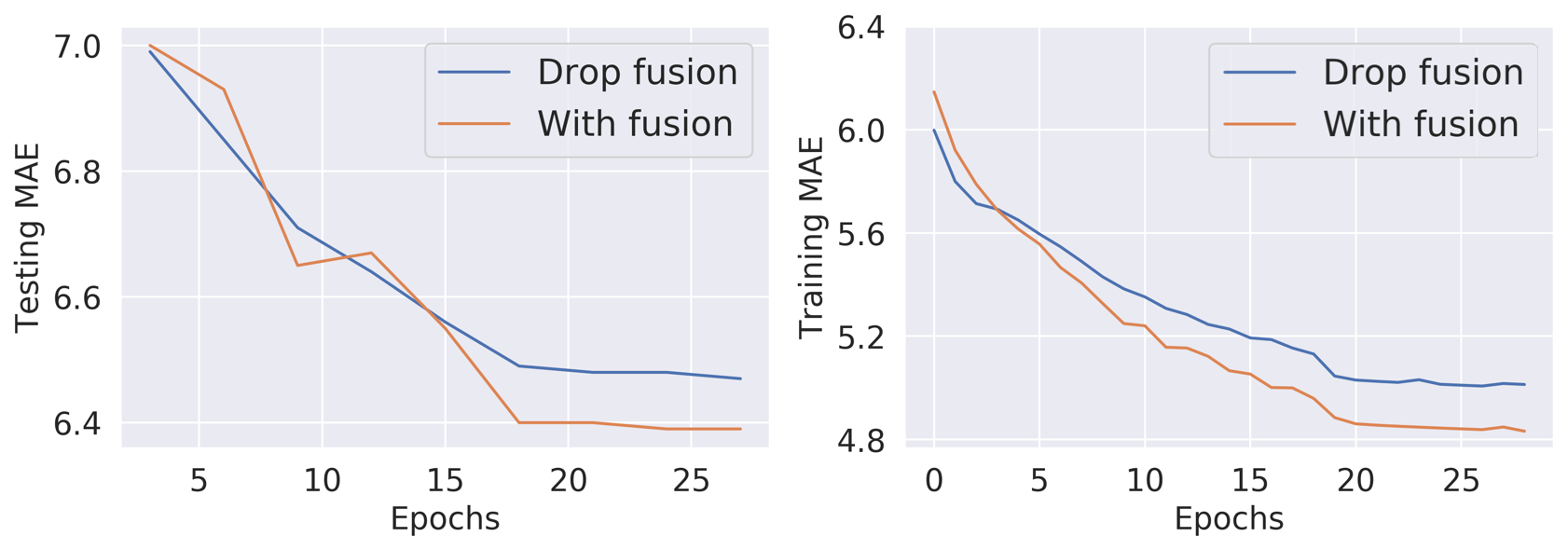}
    \caption{
    CIGNN with fusion outperforms the model without fusion, in both training and testing.}
    \label{fig:results_fusion_comp}
\end{figure}
\Paragraph{Effectiveness of Fusion}.
To assess the effectiveness of our proposed fusion mechanism, we compared our approach to a variant that removes the fusion layer.
We ran experiments on the BikeBay dataset and compared the two model for both training and testing loss, as shown in \figref{results_fusion_comp}.
Although CIGNN is initialized with a higher loss, its loss decreases faster.
This shows that the fusion mechanism is effective in utilizing contextual factors on forecasting.
We analyze the effectiveness of the Gaussian Kernel matrix and Relational matrix, added in appendix due to the limit of space.

\section{Conclusion}
To model non-linear temporal, relational, spatial, and contextual dependency in time-series predictions, we propose a novel Graph Neural Network approach for spatio-temporal data with dynamic contextual information.
Our model employs a novel fusion mechanism to capture the dynamic contextual impact on demand.

\appendix
\section{Appendix}
\begin{figure}[h]
    \centering
    \includegraphics[width=\linewidth]{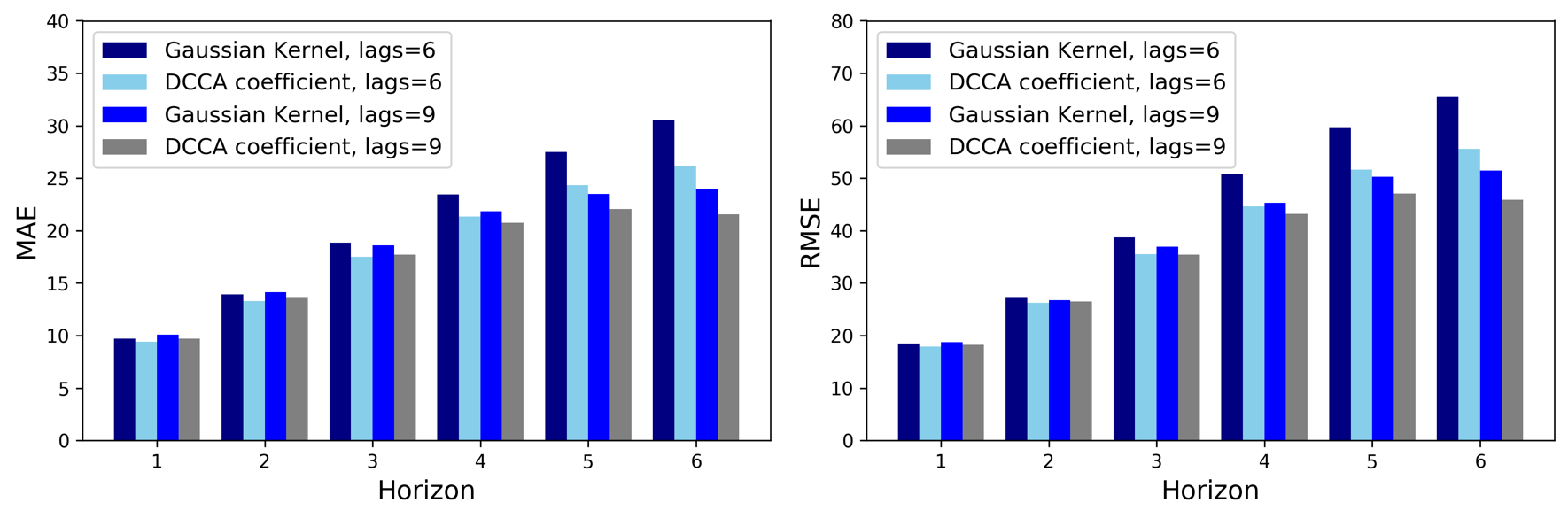}
    \caption{The MAE and RMSE comparison of models with adjacency matrix constructed by Gaussian Kernel and DCCA coefficient. 
    DCCA coefficient-based model delivers better performance in both situations when lag number is $6$ or $9$.}
    \label{fig:lags-kerDCCA-comp}
\end{figure}
\Paragraph{Adjacency Matrix Analysis.} To analyze and compare the effectiveness of the Gaussian Kernel matrix and the DCCA relational coefficients matrix, we ran experiments on the CallMi dataset using temporal lags $6$ and $9$. \figref{lags-kerDCCA-comp} shows the MAE and RMSE for a horizon of $6$. The results show that: (1) Predictions based on the DCCA coefficient adjacency matrix are consistently more accurate than predictions based on the Gaussian kernel matrix.
(2) The lag number has an impact on predicting values for a long horizon. Using $9$ lags results in significantly better prediction results than using $6$ lags. This demonstrates that CIGNN is better at learning long-term temporal dependencies.

\bibliography{aaai21}

\end{document}